\documentclass{article}

\PassOptionsToPackage{numbers, compress}{natbib}
\usepackage[preprint]{neurips_2025}

\usepackage[utf8]{inputenc} % allow utf-8 input
\usepackage[T1]{fontenc}    % use 8-bit T1 fonts
\usepackage{hyperref}       % hyperlinks
\usepackage{url}            % simple URL typesetting
\usepackage{booktabs}       % professional-quality tables
\usepackage{amsfonts}       % blackboard math symbols
\usepackage{nicefrac}       % compact symbols for 1/2, etc.
\usepackage{microtype}      % microtypography
\usepackage{xcolor}         % colors
\usepackage{graphicx}
\usepackage{hhline}
\usepackage{pifont}
\usepackage{amsmath}
\usepackage{multirow}
\usepackage{amssymb}
\usepackage{makecell}
\usepackage{marvosym}
\newcommand{\xmark}{\ding{55}}%
\newcommand{\cmark}{\ding{51}}%

\usepackage {verbatim}
\title{Video-GPT via Next Clip Diffusion
}

\author{%
  Shaobin Zhuang$^{\dagger}$ \\
  \scriptsize{Shanghai Jiao Tong University} \\
  \scriptsize{\texttt{hahahahaha@sjtu.edu.cn}} \\
  \And
  Zhipeng Huang$^{\dagger}$ \\
  \scriptsize{University of Science and Technology of China} \\
  \scriptsize{\texttt{ethanphuang@tencent.com}} \\
  \And
  Ying Zhang \\
  \scriptsize{WeChat Vision, Tencent Inc.}  \\
  \scriptsize{\texttt{yinggzhang@tencent.com}} \\
  \And
  \And
  Fangyikang Wang$^{\dagger}$ \\
  \scriptsize{Zhejiang University}  \\
  \scriptsize{\texttt{wangfangyikang@zju.edu.cn}} \\
  \And
  Canmiao Fu \\
  \scriptsize{WeChat Vision, Tencent Inc.}  \\
    \scriptsize{\texttt{kenelmfu@tencent.com}} \\
    \And
    Binxin Yang \\
  \scriptsize{WeChat Vision, Tencent Inc.}  \\
    \scriptsize{\texttt{binxinyang@tencent.com}} \\
    \And
    Chong Sun \\
  \scriptsize{WeChat Vision, Tencent Inc.}  \\
    \scriptsize{\texttt{waynecsun@tencent.com}} \\
    \And
    Chen Li \\
  \scriptsize{WeChat Vision, Tencent Inc.}  \\
    \scriptsize{\texttt{chaselli@tencent.com}} \\
    \And
    Yali Wang\textsuperscript{\Letter} \\
  \scriptsize{Shenzhen Institutes of Advanced Technology, Chinese Academy of Sciences} \\ 
  \scriptsize{Shanghai AI Laboratory} \\
\scriptsize{\texttt{yl.wang@siat.ac.cn}} \\
}

\begin{document}

\renewcommand{\thefootnote}{$\dagger$}\footnotetext{Work done as interns at WeChat Vision, Tencent Inc.}
\renewcommand{\thefootnote}{\Letter}\footnotetext{Corresponding author.}

\maketitle

\begin{abstract}
GPT has shown its remarkable success in natural language processing.
However,
the language sequence is not sufficient to describe spatial-temporal details in the visual world.
Alternatively,
the video sequence is good at capturing such details.
Motivated by this fact,
we propose a concise Video-GPT in this paper by treating video as new language for visual world modeling. %%加一两句解释
By analogy to next token prediction in GPT, %%参考conclusion
we introduce a novel next clip diffusion paradigm for pretraining Video-GPT.
Different from the previous works,
this distinct paradigm allows Video-GPT to tackle
both short-term generation and long-term prediction,
%%% 我写的这个over吗？ 因为我今天Intro大概逻辑，也差不多，你们再斟酌一下。
by autoregressively denoising the noisy clip according to the clean clips in the history.
Extensive experiments show our Video-GPT achieves the state-of-the-art performance on video prediction,
which is the key factor towards world modeling (Physics-IQ Benchmark: \textbf{Video-GPT 34.97 \textit{vs.} Kling 23.64 \textit{vs.} Wan 20.89}).
%写一些数值，惊艳的，和知名方法比较。
Moreover,
it can be well adapted on 6 mainstream video tasks in both video generation and understanding, 
showing its great generalization capacity in downstream.
The project page is at \href{https://zhuangshaobin.github.io/Video-GPT.github.io/}{https://zhuangshaobin.github.io/Video-GPT.github.io/}.

\end{abstract}

\section{Introduction}
\label{sec_introduction}

Over the past few years,
natural language processing has been mainly driven by training large language models (LLMs) from web-scale text data.
In particular,
GPT series \cite{Radford2018ImprovingLU, Radford2019LanguageMA, Brown2020LanguageMA, Achiam2023GPT4TR} have demonstrated remarkable generalization capabilities on various tasks,
which further opens the possibility towards general artificial intelligence.
However,
the language sequence is often good at expressing high-level abstractions, 
while 
it may not be sufficient to capture rich spatial-temporal details in the visual world \cite{Liu2024WorldMO,Yang2024VideoAT}. %% 你选Video as the new language 和 large World Model那两篇文章做参考文献
For example,
it is difficult to describe how to tie a knot by language.
%%% 你得用FIG1 的例子，说明，语言适合干抽象描述，学不了时空视觉细节。这要求你，FIG1 的图里，文字，以及那个视频，你好好选一选。要能表达，视频能干的事，语言干不了。
Alternatively,
the video sequence has been considered as a preferred candidate to describe such details,
since it can record visual knowledge of our dynamical world at different spatial and temporal resolutions \cite{Bai2023SequentialME,Liu2024WorldMO}. %% 你选Video as the new language 和 large World Model那两篇文章做参考文献
Hence,
there is a natural question: 
\textit{can we treat video as new language for visual world modeling?}

The recent studies have shown that,
video generation is a promising direction to achieve this goal \cite{openai2024videogenerationmodelsas,Bruce2024GenieGI,Kondratyuk2023VideoPoetAL}.
The typical design is video diffusion by adding noise and denoising gradually \cite{song2021scorebased,ho2020denoisingdiffusionprobabilisticmodels}.
Although such a paradigm has achieved significant progress \cite{BarTal2024LumiereAS,Kong2024HunyuanVideoAS,kuaishou2024klingai,Wang2025WanOA},
it often suffers from difficulty in long-term future prediction that is a critical factor of world modeling \cite{Ouyang2024FlexiFilmLV,Li2024ASO}.
To address this problem,
the autoregressive attempts have been made for long-context video modeling,
on analogy of next token prediction in LLMs \cite{Kondratyuk2023VideoPoetAL, Liu2024WorldMO, Bai2023SequentialME}. %%% Large World Model 就是一篇，你可以再找找别的。另外，你最好，实验里有些地方能和这个模型比较。因为他是个强相关的论文。
However,
compared with its diffusion counterpart,
the generation performance of this paradigm still needs to be further improved.
In order to leverage advantages from both paradigms,
several approaches combine diffusion and autogregressive modeling in a unified transformer style \cite{Xiao2024OmniGenUI,Deng2024CausalDT,Zhou2024TransfusionPT,huang2025wegenunifiedmodelinteractive, Hu2024ACDiTIA}.
But this design mainly works on the image domain,
without insightful analogy between language and video.

\begin{figure}[t]
    \centering
    % \vspace{-0.3cm}   
    \includegraphics[width=1\textwidth]{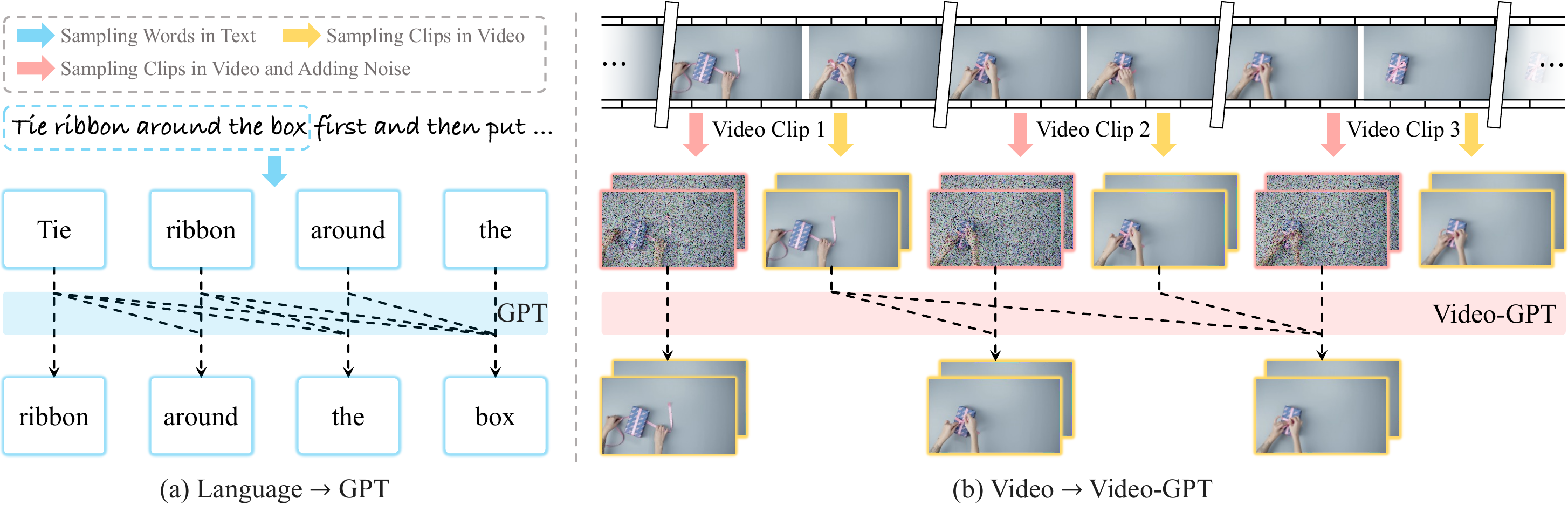}
    \vspace{-0.5cm}
    \caption{
    \textbf{Next clip diffusion.} 
    We draw an analogy with GPT's next token prediction and model each video clip as a visual word by denoising the next noisy clip, 
    conditioning on the previous video.
    %%%%在这，不该写这个。 你在这要写你的motivation。你应该写，你吧video当成新语言，模拟GPT的范式，构建 面向视觉世界建模的 Video-GPT。然后，你简要描述怎么做，就是用以前干净的CLIP，帮你把现在的噪生clip给去噪。之类之类的。在这，还没到你说training和inference的时候。
    }
    \label{fig:teaser}
    % \vspace{-0.5cm}
\end{figure}

To fill this gap,
we propose a concise Video-GPT,
%a large video model (LVM),
by analogy with generative pretraining design in GPT.
%a large language model (LLM).
Inspired by next token prediction in GPT,
we introduce a novel \textit{next clip diffusion} paradigm for our Video-GPT.
Specifically,
we creatively treat a \textit{clip} in the video as the role of a \textit{word} in the language,
since both of them describe local temporal information, 
respectively, 
in the video and language sequence.
However,
different from a discrete word token,
it is often challenging to predict a continuous video clip in the next step.
To address this difficulty,
we introduce a flexible diffusion design within an interleaved sequence of noisy and clean clips in the temporal order. 
As shown in Fig. \ref{fig:teaser},
one can randomly sample a number of clips from a training video.
For each clip,
we generate its noisy clip by adding noise on it,
based on the forward process of diffusion.
Then,
we re-arrange noisy and clean clips in the temporal order,
constructing an interleaved clip sequence.
Consequently,
for each noisy clip,
we leverage the clean clips in the previous as temporal context,
and reconstruct the corresponding clean clip by the diffusion loss.
In this case,
our Video-GPT inherits advantages from both GPT and diffusion,
but within the basic unit of a video clip.
This allows it to capture long-term video prediction as well as short-term video generation for effective generative pretraining on videos.
%%% 这句话也是一样，斟酌斟酌，over了没有。

Our contributions can be summarized as follows.
\textit{First},
we introduce a concise Video-GPT,
by analogy to GPT.
Different from GPT,
our Video-GPT aims at modeling rich spatial-temporal information in the visual world.
\textit{Second},
we design a novel next clip diffusion paradigm for pretraining Video-GPT.
Different from the previous works mentioned above,
this hybrid paradigm allows Video-GPT to tackle both short-term generation and long-term prediction,
via integrating diffusion and autoregressive design in the video clip level. 
%treating a clip as the role of word.
Finally,
our pretrained Video-GPT achieves state-of-the-art performance on Physics-IQ Benchmark and Kinetics-600,
fully demonstrating its potential for visual world modeling.
After fine-tuning on 6 downstream video generation and understanding tasks,
it also achieved preferable performance for generalization.

%贡献3 预训练的预测性能SOTA，超过各类主流开源闭源模型，显示visual World modeling的潜力，with 物理关系学习。通过fine-tune，模型可有效应用于下游视频生成与理解任务，在多少个任务上取得了优异性能。

\section{Related Work}
\label{sec_related_work}

\textbf{Video Diffusion Models.}
With diffusion models demonstrating remarkable performance in image generation \cite{Saharia2022PhotorealisticTD,Rombach2021HighResolutionIS,Dhariwal2021DiffusionMB,Ramesh2022HierarchicalTI}, 
video diffusion models \cite{Ho2022VideoDM} also surpasses traditional methods \cite{Clark2019AdversarialVG,Saito2018TrainSG,Kahembwe2019LowerDK} in video generation. 
With the emergence of large-scale text–video datasets \cite{Chen2024Panda70MC7,Wang2024Koala36MAL,Wang2023InternVidAL}, 
text-to-video diffusion models \cite{Blattmann2023StableVD,Zhuang2024VloggerMY, Yang2024CogVideoXTD, openai2024videogenerationmodelsas,Wang2025WanOA,Kong2024HunyuanVideoAS,Wang2023LAVIEHV} 
exhibit realistic generation capabilities and strong transferability to downstream tasks \cite{Yue2025VStylistVS,Hu2023VideoControlNetAM}. 
However, 
the supervised paradigm based on text annotations faces two challenges as it scales up.
\textbf{1) Quantity of Annotated Data}: 
Current methods require meticulous data filtering and annotation processes \cite{,Menapace2024SnapVS,Polyak2024MovieGA},
resulting substantially smaller data scale compared to LLM \cite{Dubey2024TheL3,Yang2024Qwen2TR,Kaplan2020ScalingLF}.
\textbf{2) Quality of Annotated Data}: 
Text cannot fully capture the meaning of a video. 
While increasing the granularity of descriptions \cite{Chen2023PixArtFT,BetkerImprovingIG,Segalis2023API} alleviates this issue, 
the difficulty of annotating such descriptions significantly increases as video length scales up.

\textbf{Autoregressive Video Models.}
ImageTransformer \cite{Parmar2018ImageT} migrated the next token prediction approach to image generation in pixel space. 
Chameleon \cite{Team2024ChameleonME} and LlamaGen \cite{Sun2024AutoregressiveMB} perform pretraining on images in latent space. 
LVM \cite{Bai2023SequentialME} and VideoWorld \cite{Ren2025VideoWorldEK} reveal that models can acquire world knowledge from videos.
LWM \cite{Liu2024WorldMO} compresses videos into latent space for pretraining, 
but focuses on video understanding capabilities. 
In general, 
next token prediction based visual generation \cite{Wang2024Emu3NP} still underperform compared with the most advanced diffusion models \cite{black2024flux,Liao2025MogaoAO,Wang2025WanOA,Kong2024HunyuanVideoAS}. 
In contrast, 
our Video-GPT adopts a next clip diffusion approach, 
which diffusion generation within video clips and applies autoregressive generation between clips. 
This method enhances video generative modeling while utilizing historical video information as a self-supervised condition.
\begin{figure}[!t]
    \centering
    % \vspace{-0.3cm}   
    \includegraphics[width=1\textwidth]{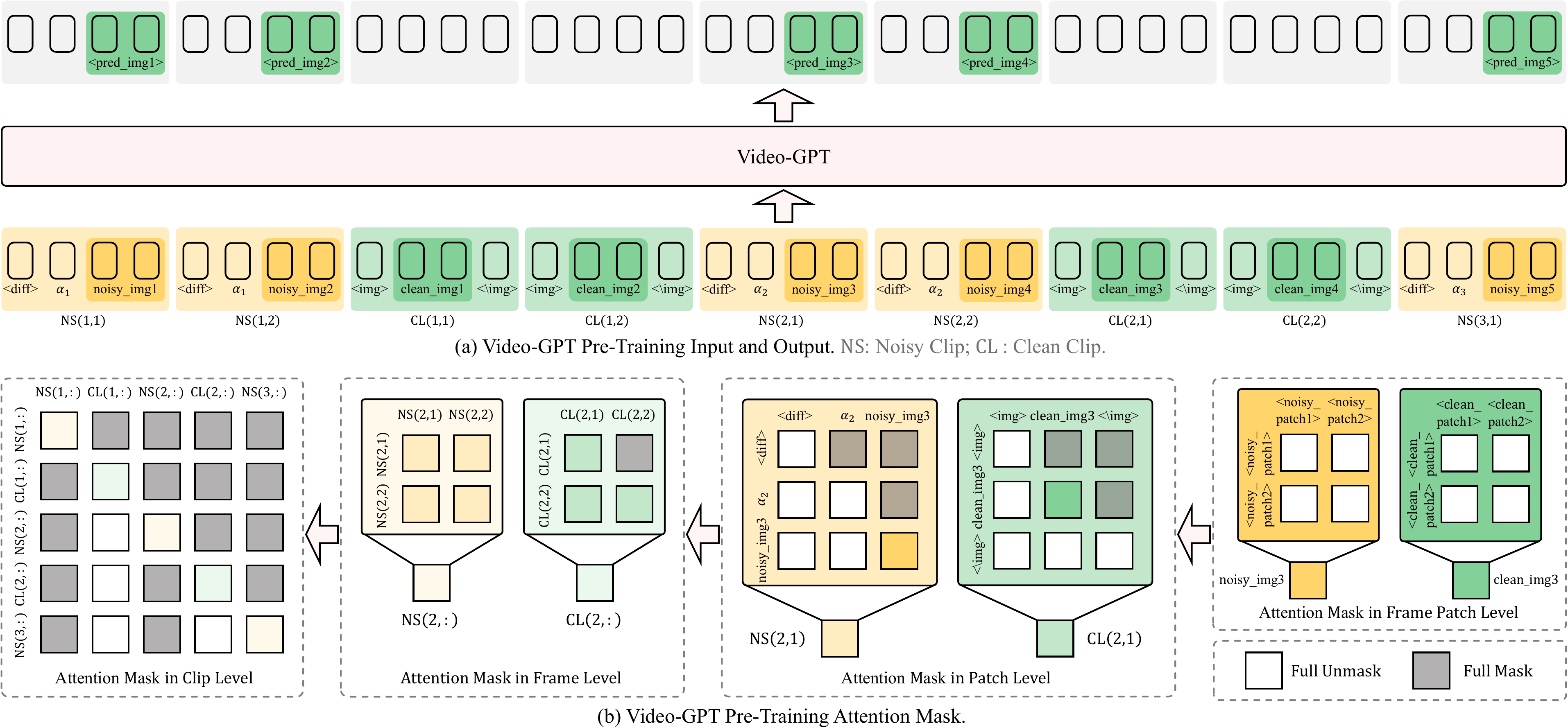}
    \vspace{-0.6cm}
    \caption{
    \textbf{Video-GPT pretraining framework.} 
    The full attention mask is shown in Fig. \ref{fig:full_mask}
    }
    \label{fig:model_pretrain}
    \vspace{-0.5cm}
\end{figure}
\begin{table*}[!t]
    \renewcommand{\arraystretch}{1}
    \centering
    \caption{\textbf{Progressive training strategy.} 
    }
    % \vspace{-0.2cm}
    \resizebox{0.9\textwidth}{!}{
    \begin{tabular}{cccccc}
        \toprule
        \textbf{Stage}
         & \textbf{Resolution}
         & \textbf{Frame Num.} 
         & \textbf{Frame Interval}
         & \textbf{Clip Num.}
         & \textbf{Train Steps} 
         \\
         \midrule
         1 & Flexible & 16 & 4 & $16$ & 300000 \\
         2 & Flexible & 48 & 4 & $\sim \text{Uniform}\{2,3,\cdots,48\}$ & 25000 \\
         3 & Flexible & 48 & $\sim \text{Uniform}\{4,5,\cdots,12\}$ & $\sim \text{Uniform}\{2,3,\cdots,48\}$ & 40000 \\
         4 & Flexible & 80 & $\sim \text{Uniform}\{4,5,\cdots,12\}$ & $\sim \text{Uniform}\{2,3,\cdots,80\}$ & 20000 \\
        \bottomrule
    \end{tabular}
    }
    \label{table: progressive_setting}
\vspace{-0.25cm}
\end{table*}

\section{Video-GPT}
\label{sec_method}

In this section, 
we first introduce our Video-GPT via next clip diffusion. 
Subsequently, 
we describe how to adapt it for typically downstream tasks in both video generation and understanding.

\subsection{Input: Clip Sequence Construction}
As mentioned before,
we notice that a clip in the video plays a similar role as a word in the language.
Hence,
we first divide a training video into a number of clips to construct the basic processing unit in our Video-GPT.
Specifically,
we uniformly sample $N$ frames from each training video.
Then,
we randomly divide these frames into $K$ clips,
where $K$ is sampled by $K \sim \text{Uniform}\{2,3,\cdots,N\}$.

\textbf{Forward Diffusion Process on Each Clip}.
Since diffusion has shown the powerful capacity of video generation,
we propose to leverage it to predict a continuous video clip in the next step.
Based on this reason,
we perform the forward diffusion process on latent of each clip by adding Gaussian noise.
Specifically,
for the $k$-th clip,
we first apply continues VAE \cite{Rombach2021HighResolutionIS} %%% 就是普通的VAE吗，还是什么，你写清楚，好像是叫VQVAE还是啥？你最好明确清楚。Vlogger上次有人质疑了你。这参考文献，我是从你底下的论文内容里摘的。你自己确认吧，他是啥，对不对。
to compress each frame in this clip,
and patchify it to obtain the latent feature of each frame
(i.e., 
the feature matrix of all the patch tokens in this frame).
Next,
we choose flow matching \cite{Lipman2022FlowMF} for diffusion, 
due to its efficiency.
%% 我写的对吗，这句。
According to flow matching,
we add the noise on the latent feature in the following way,
\begin{equation}
%\Psi^{(t)}(k,i) = \alpha(k,i)\Phi(k,i)+(1-\alpha(k,i))\varepsilon^{(t)}(k,i),
\Psi(k,i)= \alpha_{k}\Phi(k,i)+(1-\alpha_{k})\varepsilon_{k,i},
\label{eq:flow_matching_input}
\end{equation}
where
$\Phi(k,i)$ is the latent feature of the $i$-th frame in the $k$-th clip,
and 
$\Psi(k,i)$ is the noisy feature of this frame.
The weight $\alpha_{k}$ is sampled by $\alpha_{k}\sim \text{Uniform}[0,1]$,
and
the noise $\varepsilon_{k,i}$ is sampled by $\varepsilon_{k,i} \sim \mathcal{N} \left( \mathbf{0}, \mathbf{I} \right)$.
Note that,
in order to inference the content in the video clip in parallel,
we apply the same $\alpha_k$ for all frames in the video clip during training.
% \textcolor{red}{XXXXXXXXX}. 
%%% 写1句，noise和Weight，对于一个Clip里的所有帧，加的都一样。
%%% 写1句，为什么。

\textbf{Noise-Clean Interleaved Clip Sequence}.
After getting the noisy clips,
we next arrange the input sequence for our Video-GPT.
By analogy of GPT,
we leverage the historical clips as temporal context to denoise the noisy clip in the next step.
But instead of using the noisy clips in the history \cite{Chen2024DiffusionFN,Yi2025Magic1G},
we leverage the original clean clips in the history,
in order to denoise the next clip conditioned on the clean and correct context.
Hence,
we feed both noisy and clean clips as input to our Video-GPT. 
To distinguish different frames in these clips,
we add extra tokens for specializing them.
\textbf{1) Token Form of Clean Clip}: 
For each clean clip,
we add the boundary tokens with the latent feature of each frame in this clip,
\begin{equation}
\mathbf{CL}(k,i)=[\text{<img>},\Phi(k,i),\text{<\textbackslash img>}]. \label{eq:cleantoken}
\end{equation}
Hence,
the token form of the $k$-th clean clip is $\mathbf{CL}(k,:)=[\mathbf{CL}(k,1),...,\mathbf{CL}(k,N_k)]$,
where
$N_k$ is the number of frames in this clip.
\textbf{2) Token Form of Noisy Clip}: 
For each noisy clip,
we add two extra tokens with the noisy latent feature of each frame in this clip,
\begin{equation}
\mathbf{NS}(k,i)=[\text{<diff>}, \alpha_{k}, \Psi(k,i)]. \label{eq:noisytoken}
\end{equation}
where
$\text{<diff>}$ is a denoising hint token to indicate that
$\Psi(k,i)$ is noisy. %%%不是这个符号吧，文章里没有G（k,i）这个东东吧。符号写错了应该是。
Moreover,
we also add the weight $\alpha_{k}$ for the $k$-th noisy clip,
in order to provide the timestep information in the flow matching.
As a result,
the token form of the $k$-th noisy clip is $\mathbf{NS}(k,:)=[\mathbf{NS}(k,1),...,\mathbf{NS}(k,N_k)]$.
\textbf{3) Noise-Clean Interleaved Input}: 
As shown in Fig. \ref{fig:model_pretrain} (a),
after formulating both clean and noisy clips,
we arrange them together as the input sequence to Video-GPT.
Since we aim at sequential modeling,
%help with understanding and to match the order of inference, 
we take a pair of noisy and clean clips as a group,
and arrange these groups in the temporal order.
%%% 你看看我写的行不行
As a result,
we obtain an interleaved sequence of noisy and clean clips as input to our Video-GPT,
i.e.,
\begin{equation}
\mathbf{Input}=[\mathbf{NS}(1,:),\mathbf{CL}(1,:),....,\mathbf{NS}(k,:),\mathbf{CL}(k,:),...,\mathbf{NS}(K,:)]. \label{eq:input}
\end{equation}

\begin{figure}[t]
    \centering
    % \vspace{-0.3cm}   
    \includegraphics[width=1\textwidth]{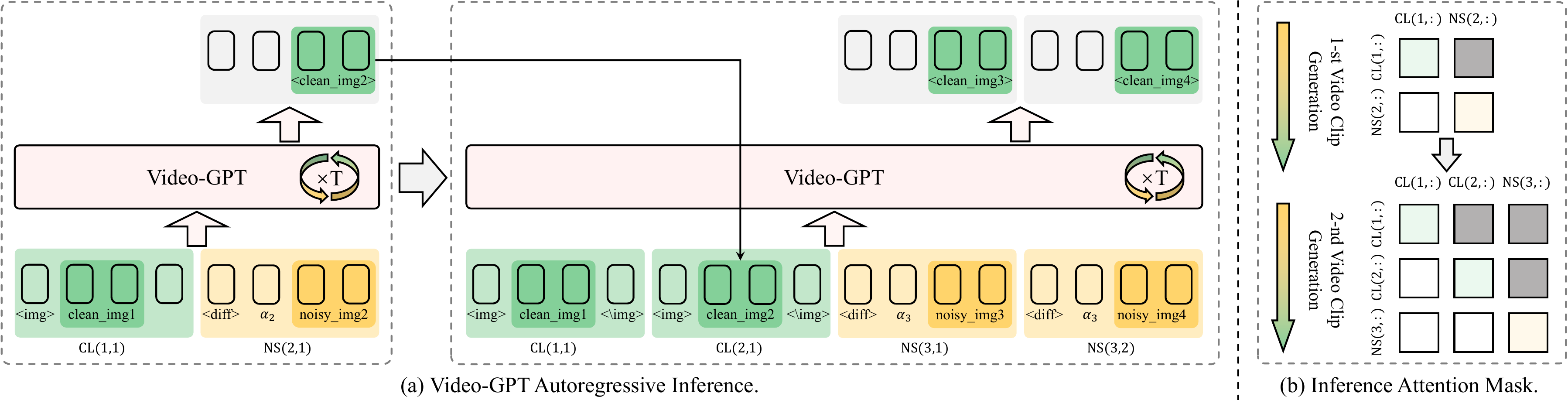}
    \vspace{-0.6cm}
    \caption{
    \textbf{Video-GPT inference framework.} 
    We iteratively denoise the 2nd noisy clip $\mathbf{NS(2,:)}$ to its clean version $\mathbf{CL(2,:)}$, 
    and use it along with the 1st clean clip $\mathbf{CL(1,:)}$ to condition the prediction of the 3rd noisy clip $\mathbf{NS(3,:)}$. 
    The number of frames in each clip can also vary during inference.
    %% 你这太简略了也，你得简要说一说怎么操作的。以自回归的形式。
    }
    \label{fig:model_inference}
    \vspace{-0.2cm}
\end{figure}

\subsection{Pretraining: Next Clip Diffusion} %%%%%%%%%%%%%%%%%%%%%%%%%%%%%%% 这一章节，一定好好的refine，多找几个同学，分别看，看看能不能读懂，有没有歧义，一定改到让所有人都能读懂的状态。%%%%%%%%%%%%%%%%%%%%%%%%%%%%%%
\label{subsec:pretraining}

After obtaining the input sequence,
we feed it into our Video-GPT for generative pretraining.
By analogy to GPT,
we use the vanilla transformer architecture for conciseness.
The next question is how to build up relations between noisy and clean clips for next clip diffusion.
As shown in Fig.~\ref{fig:model_pretrain} (b),
we introduce a hierarchical masking method to indicate such relations in the attention operation.

\textbf{Clip-Level Mask}.
As our input is a clip sequence,
we first need to define the dependence among clips.
By analogy to the dependence of words in GPT,
we basically apply a causal relation as clip dependence.
\textbf{1) Clean Clip Mask}.
As shown in Fig.~\ref{fig:model_pretrain} (b),
the $k$-th clean clip depends on itself and the previous ($k-1$) clean clips.
\textbf{2) Noisy Clip Mask}.
Alternatively,
the $k$-th noisy clip depends on itself for sure.
However,
instead of depending on the previous ($k-1$) noisy clips,
it depends on the previous ($k-1$) clean clips.
The reason is that,
these clean clips in the history provide the correct temporal context for denoising the $k$-th noisy clip.
It is the core idea in our next clip diffusion.

\textbf{Frame-Level Mask}.
Since each clip consists of several frames,
we need to further specify the dependence among frames in the defined clip-level mask.
\textbf{1) Clean Frame Mask}.
As shown in Fig.~\ref{fig:model_pretrain} (b),
for the $i$-th frame in the $k$-th clean clip,
it depends on 
itself, 
the ($i-1$) frames in this clean clip, 
and 
all the frames in the previous ($k-1$) clean clips.
\textbf{2) Noisy Frame Mask}.
Alternatively,
for the $i$-th frame in the $k$-th noisy clip,
it depends on 
itself,
all the other frames in this noisy clip, 
and 
all the frames in the previous ($k-1$) clean clips.
Note that,
different from the causal design for the clean frame,
the $i$-th noisy frame depends on all the other frames in the same noisy clip.
% instead of the ($i-1$) noisy frames in this clip.
The reason is that iterative denoising turns noisy clips into clean clip as history condition, the mask of noisy frames will not affect later inference and bidirectional attention can better enhance generation quality \cite{Yin2024FromSB,Kim2021LVerseBG}. %%%%%你读一读，这句不好读懂。不知道想说什么。
%%你要给解释，你为什么，在噪声CLIP里，每一帧，和这个CLIP里所有帧都关联。因为在干净CLIP里，每一帧，只和这个CLIP里历史帧关联。这个是比较常规的。为啥噪声的不这么设计了，要所有帧都关联，在同一个CLIP里。

\textbf{Patch-Level Mask}.
As shown in Eq. (\ref{eq:cleantoken}) and (\ref{eq:noisytoken}),
each frame (clean or noisy) is formulated by extra hint tokens,
and the latent feature of this frame which actually refers to the feature matrix of all patch tokens in this frame.
Hence,
we need to further specify the dependence among these hint tokens and patch tokens in the defined frame-level mask.
\textbf{1) Clean Patch Mask}.
As shown in Fig.~\ref{fig:model_pretrain} (b),
we follow the similar design of GPT,
the dependence is causal among tokens,
i.e.,
$\text{<img>}$, $\Phi(k,i)$, and $\text{<\textbackslash img>}$.
But differently,
the clean frame feature $\Phi(k,i)$ is actually the feature matrix of all the patch tokens in this frame.
Since the patch tokens describe spatial relation rather than temporal one,
we define the full dependence among these patch tokens.
\textbf{2) Noisy Patch Mask}.
Similar to clean patch mask,
we define the causal dependence among tokens,
i.e.,
$\text{<diff>}$, 
$\alpha_{k}$, 
and 
$\Psi(k,i)$.
We also define the full dependence among patch tokens in noisy frame as the same as the one in clean frame.
% Again,
% since the patch tokens describe spatial relation rather than temporal one,
% we define the full dependence among these patch tokens in the noisy frame.

\textbf{Training Target.}
Based on this hierarchical masking,
our Video-GPT can effectively leverage the masked attention to denoise the $k$-th noisy clip in the next step,
according to the previous $(k-1)$ clean clips of input sequence.
Subsequently,
we compute the $\mathcal{L}_2$ loss between each denoised clip feature and its corresponding clean clip feature to pretrain our Video-GPT.
It is worth mentioning that,
different from previous diffusion works,
we employ Video-GPT to predict the video clip directly instead of noise \cite{Chen2023PixArtFT,Wang2023LAVIEHV} or velocity \cite{Esser2024ScalingRF,Yang2024CogVideoXTD},
to keep the training setting as simple as possible,
allowing to easily adapt the pretrained Video-GPT for various downstream tasks.
%%%%%%%%%% 我不建议你用符号，v, e, x0啥的，这个不好理解。你就写这些东西的含义。比如，e-prediction, 应该就是 noise prediction. 你写含义，你别写符号。咋一看都是懵的。

\textbf{Progressive Training.}
As the length of the video frame increases, 
the computational cost of the attention operation will increase quadratically. 
To alleviate such computation,
we leverage a simple but effective progressive training strategy. 
Specifically,
we start training from short videos to long videos.
As shown in Tab. \ref{table: progressive_setting},
% we leverage 16 frames in the first pretraining stage and there is only one frame in each clip for next-frame prediction.
% Then,
% we progressively increase the number of clips and the number of frames in the clips for pretraining Video-GPT.
the reason is that in the initial pretraining stage we use 16 frames—with each clip containing only one frame for next-frame prediction—then gradually increase both the number of clips and frames per clip for Video-GPT pretraining.

\begin{figure}[t]
    \centering
    % \vspace{-0.3cm}   
    \includegraphics[width=1\textwidth]{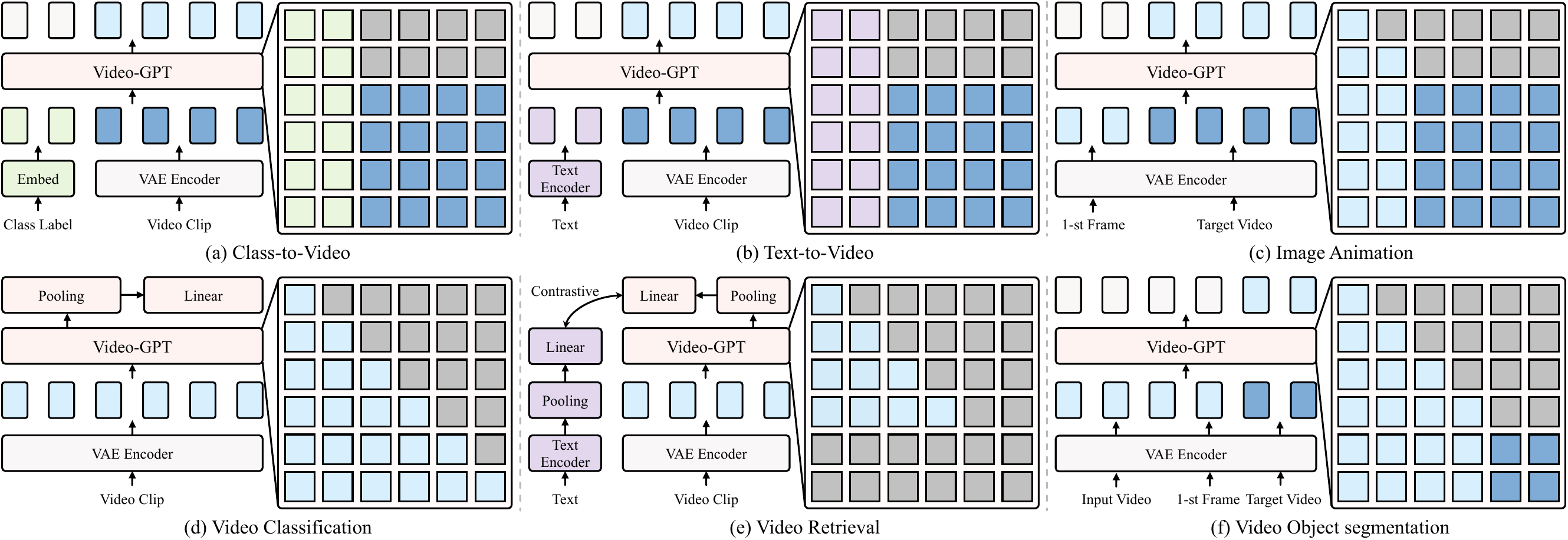}
    \vspace{-0.6cm}
    \caption{
    \textbf{Fine-tuning Video-GPT on downstream tasks.} }
    \label{fig:downstream_task}
    \vspace{-0.3cm}
\end{figure}

\subsection{Inference: Autoregressive Video Prediction}

After pretraining Video-GPT,
one can naturally leverage it for video prediction,
by analogy to GPT.
As shown in Fig. \ref{fig:model_inference} (a),
video prediction can be autoregressively performed in the inference phase.
Specifically,
we treat the previous $k$ denoised clips (outputs of Video-GPT in the history) as the clean clips,
and 
leverage them as temporal context to denoise the $(k+1)$ noisy clip,
\begin{equation}
\mathbf{DNS}(k+1,:)=\textbf{Video-GPT}\Big(~\mathbf{DNS}(1,:),...,\mathbf{DNS}(k,:),~~\mathbf{NS}(k+1,:)~\Big),
\label{eq:inference}
\end{equation}
where
$\mathbf{DNS}(k+1,:)$ is the $(k+1)$-th denoised clip.
Additionally,
$\mathbf{DNS}(1,:)$ is a clean clip which is initially given as input for video prediction.
Note that,
the input of Video-GPT in this inference phase is a causal formulation.
Hence,
the clip-level mask is defined as the causal one for each iteration of autoregressive prediction,
as shown in Fig. \ref{fig:model_inference} (b).

\subsection{Generalization on Downstream Video Tasks}
\label{subsec:method_tasks}

As we notice,
GPT has shown its powerful generalization capacity on various downstream NLP tasks.
Hence,
we expect that our Video-GPT can show the similar capacity on various downstream video tasks.
For this reason,
we adapt Video-GPT to address typical downstream video tasks in Fig. \ref{fig:downstream_task},
covering both video generation and understanding.
For video generation,
we focus on three mainstream tasks,
including 
class-to-video generation,
text-to-video generation,
and 
image animation.
For video understanding,
we focus on three mainstream tasks,
including 
video classification,
video retrieval,
and 
video object segmentation. 
%% 你最好确认一下名字，我印象中这个不是tracking，好像叫 Video Object Segmentation 之类的。

\begin{figure}[t]
    \centering
    % \vspace{-0.3cm}   
    \includegraphics[width=0.9\textwidth]{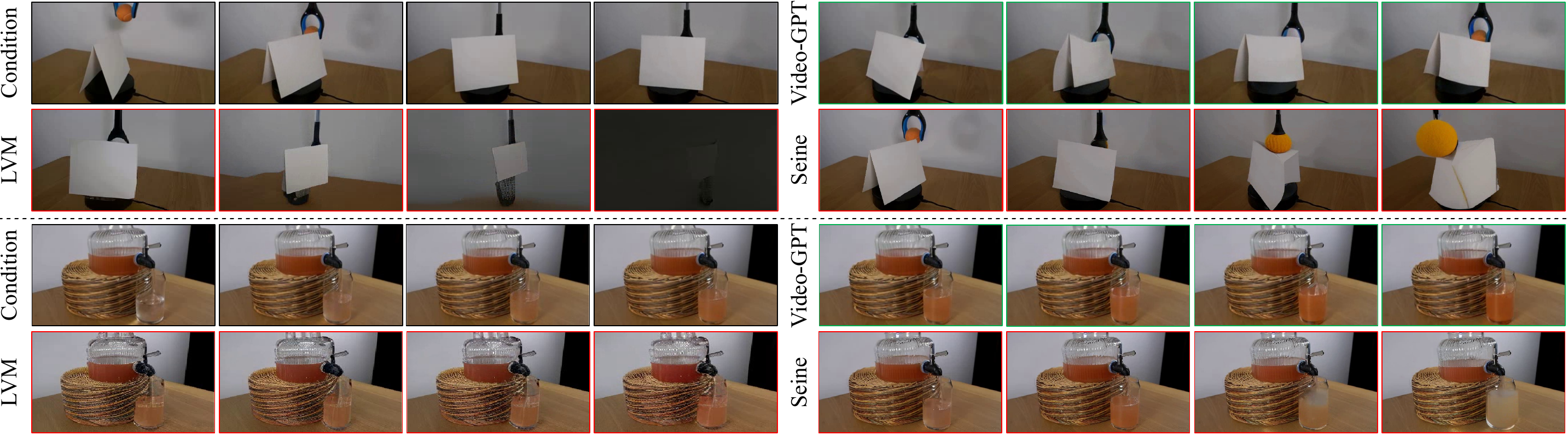}
    \vspace{-0.25cm}
    \caption{
    \textbf{Qualitative results on Physics-IQ Benchmark.} 
    The videos predicted by our Video-GPT based on condition frames are more consistent with physical laws than other methods \cite{Chen2023SEINESV, Bai2023SequentialME}.
    }
    \label{fig:phy_bench_vis}
    \vspace{-0.55cm}
\end{figure}

\begin{table*}[!t]
    \renewcommand{\arraystretch}{1}
    \centering
    % \vspace{-0.15cm}
    \caption{\textbf{Quantitative comparison of models evaluated on the Physics-IQ Benchmark.}} %%%%I2V和V2V分开写
    \resizebox{0.9\textwidth}{!}{
    \begin{tabular}{l|c|cccc|c}
        \toprule
        \multirow{2}{*}{\textbf{Model}} & \multicolumn{1}{c|}{\textbf{\textit{w/o.}}} & \multicolumn{1}{c}{\textbf{Spatial}} & \multicolumn{1}{c}{\textbf{Spatio}} & \multicolumn{1}{c}{\textbf{Weighted}} & \multirow{2}{*}{\textbf{MSE↓}} & \multicolumn{1}{c}{\textbf{Phys.}} \\
        ~ & \multicolumn{1}{c|}{\textbf{Text}} & \multicolumn{1}{c}{\textbf{IoU↑}} & \multicolumn{1}{c}{\textbf{Temporal↑}} & \multicolumn{1}{c}{\textbf{Spatial IoU↑}} & ~ & \multicolumn{1}{c}{\textbf{IQ Score↑}} \\
        \midrule
        Sora (I2V) \cite{openai2024videogenerationmodelsas} & \textcolor[RGB]{234,52,35}{\xmark} & 0.138 & 0.047 & 0.063 & 0.030 & 10.00\\
        Pika 1.0 (I2V) \cite{pikalabs2024pika} & \textcolor[RGB]{234,52,35}{\xmark} & 0.140 & 0.041 & 0.078 & 0.014 & 13.00 \\
        SVD (I2V) \cite{Blattmann2023StableVD} & \textcolor[RGB]{79,173,91}{\cmark} & 0.132 & 0.076 & 0.073 & 0.021 & 14.80 \\
        LVM (V2V) \cite{Bai2023SequentialME} & \textcolor[RGB]{79,173,91}{\cmark} & 0.100 & 0.147 & 0.077 & 0.021 & 18.02 \\
        Lumiere (I2V) \cite{BarTal2024LumiereAS} & \textcolor[RGB]{234,52,35}{\xmark} & 0.113 & 0.173 & 0.061 & 0.016 & 19.00 \\
        Open-Sora-Plan v1.3.0 (I2V) \cite{Lin2024OpenSoraPO} & \textcolor[RGB]{234,52,35}{\xmark} & 0.142 & 0.139 & 0.074 & 0.021 & 19.42 \\
        Wan2.1 (I2V) \cite{Wang2025WanOA} & \textcolor[RGB]{234,52,35}{\xmark} & 0.153 & 0.100 & 0.112 & 0.023 & 20.89 \\
        Gen 3 (I2V) \cite{runway2024gen3} & \textcolor[RGB]{234,52,35}{\xmark} & 0.201 & 0.115 & 0.116 & 0.015 & 22.80 \\
        Lumiere (V2V) \cite{BarTal2024LumiereAS} & \textcolor[RGB]{234,52,35}{\xmark} & 0.170 & 0.155 & 0.093 & 0.013 & 23.00 \\
        Kling1.6 (I2V) \cite{kuaishou2024klingai} & \textcolor[RGB]{234,52,35}{\xmark} & 0.197 & 0.086 & 0.144 & 0.025 & 23.64 \\
        Seine (V2V) \cite{Chen2023SEINESV} & \textcolor[RGB]{234,52,35}{\xmark} & 0.163 & 0.208 & 0.131 & 0.012 & 29.13 \\
        VideoPoet (V2V) \cite{Kondratyuk2023VideoPoetAL} & \textcolor[RGB]{234,52,35}{\xmark} & \textbf{0.204} & 0.164 & 0.137 & 0.010 & 29.50 \\
        \midrule
        \textbf{Video-GPT} (V2V) & \textcolor[RGB]{79,173,91}{\cmark} & 0.198 & \textbf{0.240} & \textbf{0.144} & \textbf{0.007} & \textbf{34.97} \\
        \bottomrule
    \end{tabular}}
    \label{table:physics_iq}
    \vspace{-0.55cm}
\end{table*}
\begin{table*}[!t]
    \renewcommand{\arraystretch}{1}
    \centering
    \caption{\textbf{Quantitative comparison of video generation models evaluated on the Kinetics-600.} }
    % \vspace{-0.12cm}
    \resizebox{0.9\textwidth}{!}{
    \begin{tabular}{l|c|c|c|c}
        \toprule
        \textbf{Model} & \textbf{Parameters} & \textbf{Architecture} & \textbf{Text Encoder} & \textbf{FVD↓}\\
        \midrule
        Open-Sora-Plan v1.3.0 \cite{Lin2024OpenSoraPO} & 2B & DiT & mT5-XXL \cite{Xue2020mT5AM} & 384.27 \\
        LVM \cite{Bai2023SequentialME} & 7B & Vanilla Transformer & - & 356.48 \\
        Seine \cite{Chen2023SEINESV} & 0.9B & U-Net & CLIP-ViT-L \cite{Radford2021LearningTV} & 332.80 \\
        \midrule
        \textbf{Video-GPT} & 3.8B & Vanilla Transformer & - & \textbf{315.40} \\
        \bottomrule
    \end{tabular}}
    \label{table:k600}
    \vspace{-0.2cm}
\end{table*}

\textbf{Class-to-Video Generation and Text-to-Video Generation.}
As shown in Fig. \ref{fig:downstream_task},
we use a similar processing mode for these two types of tasks,
%%%%%%%%%%% 你FIG4里的所有Video-GPT都得改成 Video-GPT
since both generation tasks of Video-GPT are conditioned on extra tokens (either from a category tag or a video description).
Hence,
we introduce a causal attention between extra tokens and clip tokens,
i.e.,
a extra token only depends on other extra tokens,
while
a clip token depends on both extra and clip tokens.

%We convert the category label or text description of the video $V$ into condition tokens $c$ through the embedding layer or text encoder. 
%For the video $V$, 
%we encode it through VAE and add noise according to Eq. \ref{eq:flow_matching_input} to get $v'$, 
%which is then concatenated with $c$ to get the model input.

%Referring to the process of fine-tuning LLM to VLM, 
%we use a causal attention mechanism between $c$ and $v'$, 
%that is, 
%the token in $c$ can only pay attention to other tokens in $c$, 
%while the token in $v'$ can pay attention to all tokens.

\textbf{Image Animation and Video Object Segmentation}. %% 你记得统一改名字，如果要是名字不对的话。我底下就不一一改了。
Since both tasks are to predict video content conditioned on the given sequence,
we discuss them together here.
For image animation,
we are given the first frame.
The goal is to predict the rest video conditioned on the first frame. 
%%%%%%%%% 你检查一些，你写的对不对，这个任务没有文本？ 你看你后面给的可视化，这个任务是有文本的，比如 lie down, transform and fly. 你要是弄错了，你这个文字里，以及你的FIG4，都要改。
For video object segmentation,
we are given a video and the first frame with the masked objects.
The goal is to track the masked objects throughout the video.
To address these tasks by fine-tuning,
we leverage the input frame or video as the clean clip in our Video-GPT,
and 
the video to be generated as the noisy clip.
The mask strategy follows the settings in Sec. \ref{subsec:pretraining}.

%The difference between them lies in the content and number of given frames.
%Image animation only gives the first frame of the video in a specific domain and predicts the content of the rest of the video, 
%while video object tracking gives a video and the first frame of video with masked, 
%and then lets the model continue to track the masked part of the first frame.
%We use a method similar to the Supervised Fine-Tuning (SFT) in LLM to train our Video-GPT,

%we regard the given frames as an instruction and use the given frames as clean video clip $v$. 
%During the training process, 
%only the content to be generated is used as noised video clip $v'$ for supervised learning.
%The mask strategy follows the settings in Sec. \ref{subsec:method_model}.

\textbf{Video Classification and Video Retrieval.}
% For video classification,
% we extract the clip features from Video-GPT and pool them for linear probing.
% For video retrieval,
% we use the same operation to obtain video feature,
% and 
% perform contrastive learning between video feature and text feature extracted from text encoder.
For video classification, 
we pool clean clip features from Video-GPT for linear probing.
For video retrieval, 
we use the same clean clip feature for contrastive learning with the text feature from a text encoder, 
e.g., 
CLIP-ViT-Large-Patch14 \cite{Radford2021LearningTV}. %你用的text encoder是谁，你再在说一说，我没看见别的地方有这个。
%%%%%%%%%%% Fig4 的retrieval画的对吗，怎么前两列，都是黑色的Token，他应该和Video classification那个画的一样吧。

%We use Video-GPT as a video encoder, 
%and videos are input into Video-GPT as clean video clips. 
%And we use the same mask as described in Sec. \ref{subsec:method_model}. 
%In video retrieval, 
%we also perform comparative learning based on a pretrained language model to make the model converge to better performance.

\section{Experiments}
\label{sec_experiments}

\subsection{Implementation}
\label{subsec:imple}
\textbf{Dataset.}
In pretraining, 
we use videos in Panda-70M \cite{Chen2024Panda70MC7} as the training dataset. 
In order to verify the impact of data scale on model performance, 
we also use videos in OpenVid-1M \cite{Nan2024OpenVid1MAL} as the training dataset in Sec. \ref{subsec:ablation_study}.
For downstream tasks, 
we fine-tuned the model using supervised datasets. 
Specifically, 
for class-to-video and video classification, 
we conduct experiments on UCF-101 \cite{Soomro2012UCF101AD}. 
For text-to-video generation, 
we fine-tune the model using videos and annotated texts from the OpenVid-1M, 
and for the video retrieval task, 
we used videos and annotated texts from the Panda-70M. 
For video object segmentation, 
we randomly select 1000 cases from GetIn-1M \cite{Zhuang2025GetIV}.
Notably, 
for image animation—which is in high demand within the industry,
there is no commonly recognized benchmark. 
Therefore, 
we colloct three datasets from the internet to evaluate the generalization capability of pretrained Video-GPT on image animation. 
These datasets correspond to the categories of 
\textit{lie down},
\textit{transform}, 
and \textit{fly}, 
with each dataset containing fewer than one hundred videos.

\begin{table*}[!t]
    \renewcommand{\arraystretch}{1}
    \centering
    % \vspace{-0.15cm}
    \caption{\textbf{Inference setting ablation.}} 
    \resizebox{0.95\textwidth}{!}{
    \begin{tabular}{l|cccccc|cccccc}
        \toprule
        ~ & \multicolumn{6}{c|}{\textbf{Number of Frames in Each Video Clip}} & \multicolumn{6}{c}{\textbf{History Conditioned Classifier-Free Guidance}} \\
        \midrule
        \textbf{Hyperparams.} & 1 & 2 & 4 & 8 & 16 & \textbf{32} & 1 & 1.5 & 2.0 & 2.5 & \textbf{3.0} & 3.5 \\
        \midrule
        \textbf{Phy. IQ Score} & 0.00 & 6.34 & 17.42 & 28.54 & 32.72 & \textbf{33.09} & 25.09 & 31.96 & 32.22 & 32.65 & \textbf{32.72} & 31.88 \\
        \bottomrule
    \end{tabular}}
    \label{table:abla_infer}
    \vspace{-0.6cm}
\end{table*}
\begin{table*}[!t]
    \renewcommand{\arraystretch}{1}
    \centering
    % \vspace{-0.15cm}
    \caption{\textbf{Training setting ablation.}} 
    \resizebox{0.95\textwidth}{!}{
    \begin{tabular}{l | c c c | c c | c c }
        \toprule
        ~ & \multicolumn{3}{c|}{\textbf{Number of Frames Sampled from Origin Video}} & \multicolumn{2}{c|}{\textbf{Whether to Add Noise to the Clean Clip}} & \multicolumn{2}{c}{\textbf{Scale of Dataset}} \\
        \midrule
        \textbf{Hyperparams.} & \quad\quad 16 & \quad\quad\quad 48 &  \textbf{80} & \quad\quad\quad\quad NO & \quad \textbf{YES} & 1M & \textbf{70M} \\
        \midrule
        \textbf{Phy. IQ Score} & \quad\quad 22.06 & \quad\quad\quad 33.09 &  \textbf{34.94} & \quad\quad\quad\quad 32.54 & 
\quad \textbf{33.09} & 23.16 & \textbf{33.09} \\
        \bottomrule
    \end{tabular}}
    \label{table:abla_train}
    \vspace{-0.3cm}
\end{table*}

\textbf{Settings.}
In this paper, 
we utilize VAE from SDXL \cite{Podell2023SDXLIL}, 
with Video-GPT inheriting the model architecture from Phi-3-mini \cite{Abdin2024Phi3TR}, 
comprising 3.8B parameters. 
We additionally incorporated \texttt{clean\_input} and \texttt{noised\_input} layers \cite{Xiao2024OmniGenUI} to transform VAE-encoded image latents into tokens, 
as well as a \texttt{noised\_output} layer to convert output tokens back into image latents, 
with these components having significantly fewer parameters than the main transformer body. 
During the pretraining phase, 
we employed 320 A100 GPUs and implemented a progressive training strategy as shown in Tab. \ref{table: progressive_setting}. 
% To extend the video duration processed during pretraining, 
% we gradually increased the number of training frames from 16 to 80 across training stages, 
% while also expanding the Frame Interval from 4 to 12. 
% This means that for a typical internet video at 24 FPS, 
% Video-GPT's pretraining window can reach up to 40 seconds.
% To better accommodate videos of varying resolutions, 
% we consistently applied a dynamic resolution strategy (resizing the longest edge of each frame to 320 pixels) throughout Video-GPT training, 
% and used padding to address token length mismatches between different batches caused by resolution differences. 
% We implemente the AdamW \cite{Loshchilov2017DecoupledWD} optimizer, set the learning rate to 1e-4 and perform a 1000 step warm up.
% We utilize bf16 mixed precision training, 
% FlashAttention \cite{Dao2022FlashAttentionFA}  and DeepSpeed ZeRO-2 \cite{Rasley2020DeepSpeedSO} to optimize memory usage.
More pretraining and fine-tuning settings can be found in Sup. \ref{sup:implementation}.

\subsection{Pretrained Model Evaluation}
We evaluated Video-GPT pretraining performance on Physics-IQ Benchmark \cite{motamed2025generativevideomodelsunderstand} and Kinetics-600 \cite{Carreira2018ASN}, 
which are used to evaluate the ability of pretrained models in predicting highly deterministic and highly uncertain videos respectively.
We set the historical conditioned classifier-free guidance scale $c=3$, 
and all subsequent experiments will keep the same setting if there is no callback.

\textbf{Physics-IQ Benchmark.}
% As one of the most representative benchmarks for evaluating a model’s ability to capture physical dynamics in video, 
Physics-IQ Benchmark presents a short video clip (3 seconds) depicting real-world physical motion and asks the model to predict future frames (5 seconds). 
Since macroscopic physical laws typically exhibit deterministic behavior, 
we used the Physics-IQ Benchmark to assess the model's ability to predict highly deterministic future events.
As shown in Tab. \ref{table:physics_iq},
Video-GPT has a far superior performance in modeling the physical world (more than \textbf{5 points} higher than the second place). 
In addition to our Video-GPT, 
the results of LVM \cite{Bai2023SequentialME}, 
Open-Sora-Plan v1.3 \cite{Lin2024OpenSoraPO}, 
and Seine \cite{Chen2023SEINESV} are also obtained by our own testing. 
This result shows that the model based on our next clip diffusion pretraining paradigm can better learn world knowledge from video data.

\textbf{Kinetics-600.}
Kinetics-600 contains 0.3M human motion videos,
and human motion is often known for being unpredictable and diverse.
Therefore,
we measure the distance \cite{Unterthiner2018TowardsAG} between model-generated videos and videos from Kinetics-600 to evaluate the model's capability to predict highly uncertain future events, 
requiring the model to predict 13 future frames based on the first 3 given frames. 
We randomly sampled 500 videos from Kinetics-600.
As shown in Tab. \ref{table:k600}, 
Video-GPT achieves the best FVD among all models with a vanilla transformer architecture, 
rather than the more popular U-Net \cite{Ronneberger2015UNetCN} or DiT \cite{Peebles2022ScalableDM},
which fully demonstrates the effectiveness of our pretraining.

\textbf{Visualization.}
As shown in Fig. \ref{fig:phy_bench_vis},
Video-GPT accurately predicts the water filling and machine lifting physics,
while other methods \cite{Bai2023SequentialME,Chen2023SEINESV} suffer from mode collapse or prediction errors.

\subsection{Ablation Study}
\label{subsec:ablation_study}
We ablate the properties of Video-GPT pretraining and inference in the Physics-IQ Benchmark \cite{motamed2025generativevideomodelsunderstand}.
Unless specified,
we perform ablation experiments based on the Stage 3 pretraining model in Tab. \ref{table: progressive_setting}.

\begin{figure}[t]
    \centering
    % \vspace{-0.3cm}   
    \includegraphics[width=1\textwidth]{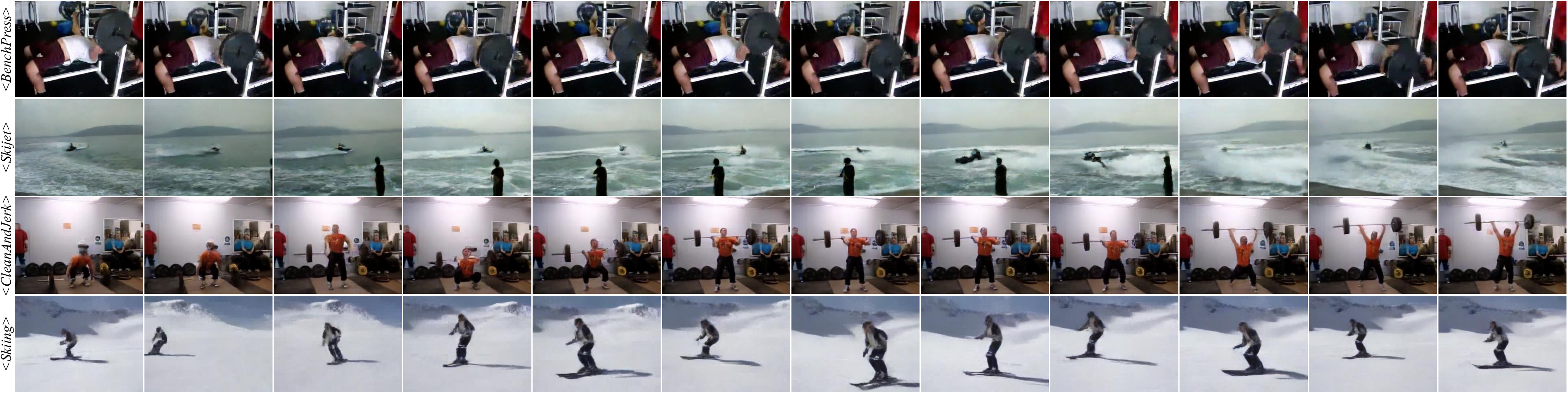}
    \vspace{-0.65cm}
    \caption{
    \textbf{Qualitative results of Video-GPT on class-to-video generation on UCF-101.} 
    % Pretrained or Lora-tuned model typically generates images that are not aligned with the prompt, while TSM benefits visual quality and text alignment.
    }
    \label{fig:ucf101_gen}
    \vspace{-0.55cm}
\end{figure}

\begin{table*}[!t]
    \centering
    % 左侧：第一个表（整个表宽度设为55%）
    \begin{minipage}[t]{0.792\linewidth}
        \vspace{0pt}
        \renewcommand{\arraystretch}{1}
        \caption{\textbf{Class to video quantitative comparison on the UCF-101.}}
        \resizebox{1\linewidth}{!}{%
        \begin{tabular}{l|c|c|c|c}
            \toprule
            \textbf{Model} & \textbf{Resolution} & \textbf{Architecture} & \textbf{VAE} & \textbf{FVD↓}\\
            \midrule
            CogVideo \cite{Hong2022CogVideoLP} & 160 $\times$ 160 & Dual-channel Transformer & 2D & 626 \\
            Latte \cite{Ma2024LatteLD} & 256 $\times$ 256 & DiT & 2D & 478 \\
            TATS \cite{Ge2022LongVG} & 128 $\times$ 128 & Time-Sensitive Transformer & - & 332 \\
            OmniTokenizer  \cite{Wang2024OmniTokenizerAJ} & 256 $\times$ 256 & Vanilla Transformer & 3D & 191 \\
            VideoFusion  \cite{Luo2023VideoFusionDD} & 128 $\times$ 128 & U-Net & - & 173 \\
            MAGVITv2-AR  \cite{Yu2024LanguageMB} & - & Vanilla Transformer & 3D & 109 \\
            ACDIT \cite{Hu2024ACDiTIA} & - & DiT & 2D & 90 \\
            Make-A-Video \cite{Singer2022MakeAVideoTG} & 256 $\times$ 256 & U-Net & 2D & 81 \\
            MAGVITv2  \cite{Yu2024LanguageMB} & - & Vanilla Transformer & 3D & 58 \\
            LARP  \cite{Wang2024LARPTV} & 128 $\times$ 128 & Vanilla Transformer & 3D & 57 \\
            FAR  \cite{Gu2025LongContextAV} & 128 $\times$ 128 & DiT & 2D & 57 \\
            \midrule
            Video-GPT (from scratch) & 240 $\times$ 320 & Vanilla Transformer & 2D & 489 \\
            \textbf{Video-GPT} (finetune) & \textbf{240 $\times$ 320} & \textbf{Vanilla Transformer} & \textbf{2D} & \textbf{53} \\
            \bottomrule
        \end{tabular}%
        }
        % \vspace{-0.5pt}
        \label{table:c2v}
    \end{minipage}
    \hfill
    % 右侧：右侧整体 minipage（宽度40%），内部上下排列两个表格
    \begin{minipage}[t]{0.202\linewidth}
        \centering
        \vspace{0pt}
        % 第二个表 --- 视频分类线性探测表（位于上面）
        \caption{Video classification linear probe on UCF-101.}
        % \vspace{-0.3cm}
        \resizebox{\linewidth}{!}{%
        \begin{tabular}{l|c}
            \toprule
            \textbf{Method} & Top-1 \\
            \midrule
            MemDPC \cite{Han2020MemoryaugmentedDP} & 54.1 \\
            VideoMAEv2 \cite{Wang2023VideoMAEVS} & 56.4 \\
            \midrule
            \textbf{Video-GPT} & \textbf{58.9} \\
            \bottomrule
        \end{tabular}%
        }
        
        \label{tab:video_classification}
        
        \vspace{-0.06cm} % 调整两个表间的垂直间距
        
        % 第三个表 --- 视频检索零样本表（位于下方）
        % \vspace{-0.15cm}
        \caption{Video retrieval zero-shot on the MSR-VTT.}
        \resizebox{0.9\linewidth}{!}{%
        \begin{tabular}{l|c}
            \toprule
            \textbf{Method} & R@1 \\
            \midrule
            VideoCLIP \cite{Xu2021VideoCLIPCP} & 10.4 \\
            SupportSet \cite{Patrick2020SupportsetBF} & 12.7 \\
            FiT \cite{Bain2021FrozenIT} & 18.8 \\
            \midrule
            \textbf{Video-GPT} & \textbf{22.8} \\
            \bottomrule
        \end{tabular}%
        }
        \label{tab:video_retrieval}
    \end{minipage}
\end{table*}

\begin{figure}[t]
    \centering
    % \vspace{-0.3cm}   
    \includegraphics[width=1\textwidth]{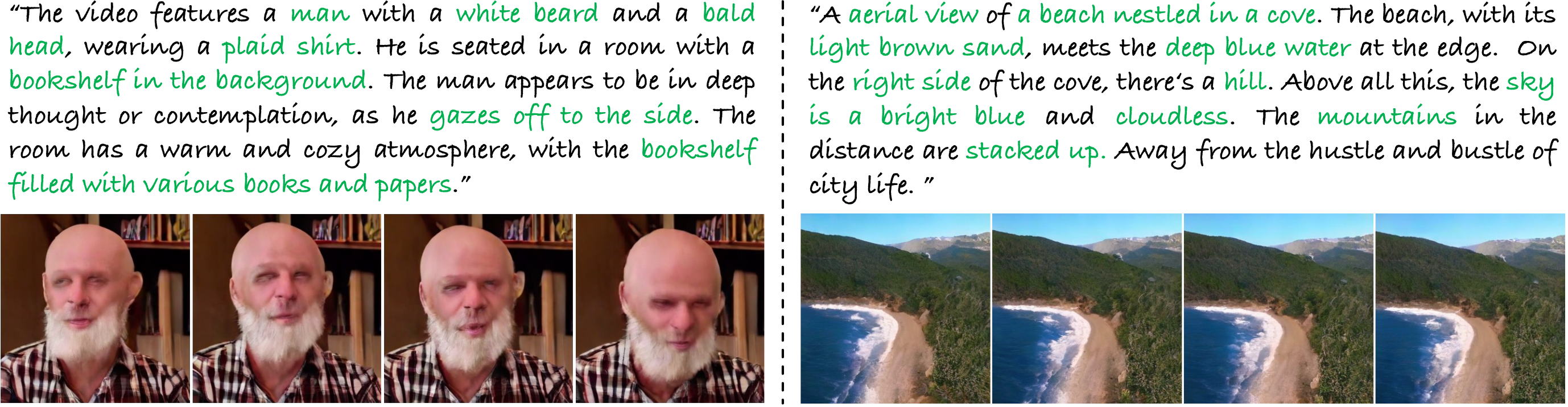}
    \vspace{-0.65cm}
    \caption{
    \textbf{Qualitative results of Video-GPT on text-to video-generation.} 
    }
    \label{fig:T2V}
    \vspace{-0.35cm}
\end{figure}

\begin{figure}[t]
    \centering
    % \vspace{-0.3cm}   
    \includegraphics[width=1\textwidth]{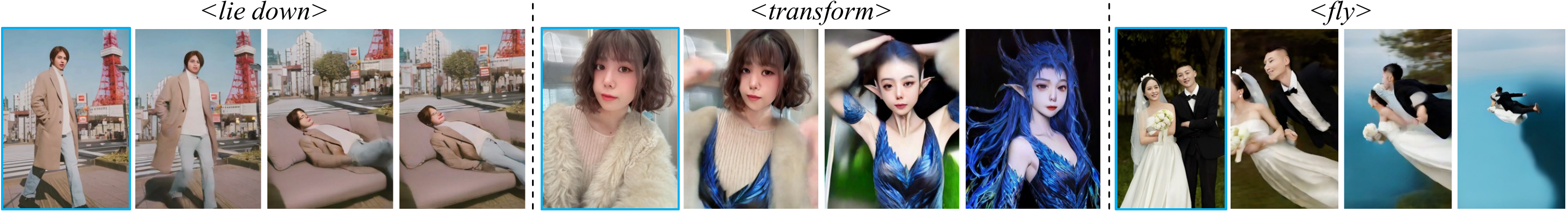}
    \vspace{-0.6cm}
    \caption{
    \textbf{Qualitative results of Video-GPT on \textit{lie down}, \textit{transform} and \textit{fly} image animation.
    } 
    The frame in the \textcolor[RGB]{79,173,234}{blue} box is the prefix frame we are given as history condition.
    }
    \label{fig:image_animation}
    \vspace{-0.3cm}
\end{figure}

\begin{figure}[!htbp]
    \centering
    % \vspace{-0.3cm}   
    \includegraphics[width=1\textwidth]{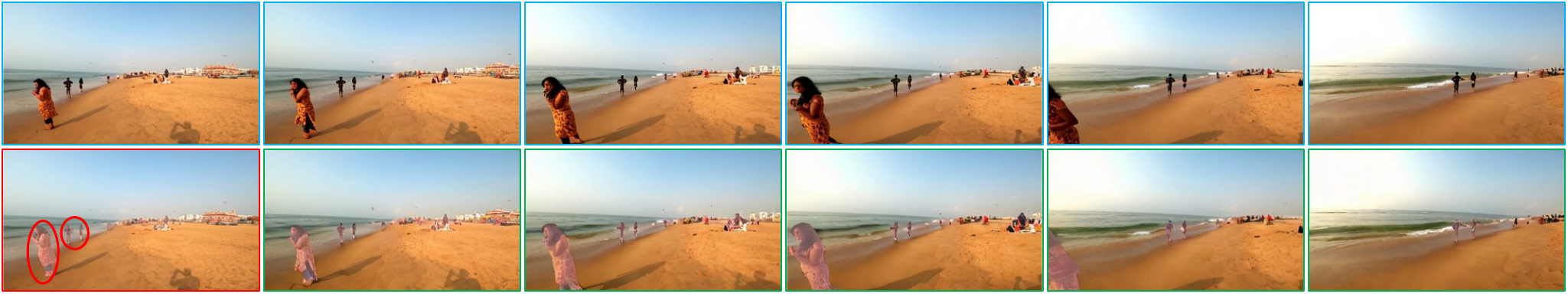}
    \vspace{-0.5cm}
    \caption{
    \textbf{Qualitative results of Video-GPT on video object segmentation.
    } 
    The frames in \textcolor[RGB]{79,173,234}{blue} box are condition frames.
    The frame in \textcolor[RGB]{234,52,35}{red} box is the 1-st frame of the condition with object mask,
    and we circle the segmented object in \textcolor[RGB]{234,52,35}{red}.
    Frames in \textcolor[RGB]{79,173,91}{green} box are the generated segmentation result.
    }
    \label{fig:video_object_seg}
    % \vspace{-0.1cm}
    %%%%% 试图测量数值结果
\end{figure}

\textbf{Number of Frames in Video Clip for Inference.}
During inference, 
we set the $N_k$ for each inference to be the same, 
that is, 
$N_1$$=$$\cdots$$=$$N_k$$=$$\cdots$$=$$N_K$.
As shown in Tab. \ref{table:abla_infer},
within a certain range, 
as the number of frames $N_k$ processed in parallel for inference within each video clip increases, 
the quality of the videos generated by the model improves significantly. 
This thoroughly validates the superiority of our proposed next clip generation paradigm for video generation tasks.

\textbf{History Conditioned Classifier-Free Guidance Scale for Inference.}
We conduct experiments based on $N_k$$=$$16$.
As shown in Tab. \ref{table:abla_infer}, 
the model shows the strongest performance
when we adjust the scale of history conditioned classifier-free guidance \cite{Ho2022ClassifierFreeDG} $c$$=$$3.0$.

\textbf{Number of
Frames for PreTraining.}
As shown in Tab. \ref{table:abla_train}, 
the results show that as the number of pretraining frames increases, 
the videos generated by Video-GPT are more and more in line with physical laws. 
This reveals that as the time window of model pretraining increases, 
the model's modeling of world knowledge is getting better and better.

\textbf{Add Noise to Clean Clip for Pretraining.}
During inference, 
there is a deviation between $\mathbf{DNS}$ and $\mathbf{CL}$.
To fill this deviation,
we add slight noise to clean frames during training \cite{Wu2022DenoisingMA,Zhou2025TamingTF} as
\begin{equation}
\Phi(k,i) = (\beta+\gamma_{k,i})\Phi(k,i)+(1-\beta-\gamma_{k,i})\epsilon_{k,i},
\label{eq:input_noise}
\end{equation}
where $\beta$ and $\gamma_{k,i}\sim\text{Uniform[0,1]}$ refer to the basic retention degree and random retention degree of the clean frame respectively and $\epsilon_{k,i}$ is sampled by $\epsilon_{k,i}\sim \mathcal{N} \left( \mathbf{0}, \mathbf{I} \right)$.
We set $\beta$$=$$0.9$. 
As shown in Tab. \ref{table:abla_train},
the result shows that adding noise to the clean clip during training improves model performance.

\textbf{Dataset Scale for Pretraining.}
We conducted ablation studies on the impact of dataset scale based on OpenVid-1M \cite{Nan2024OpenVid1MAL} and Panda-70M \cite{Chen2024Panda70MC7}. 
Considering that the videos in OpenVid-1M are a subset of Panda-70M, 
our ablation experiments are conducted under approximately identical data distributions. 
As shown in Tab. \ref{table:abla_train}, 
the results demonstrate that after expanding the pretraining video dataset scale, 
Video-GPT's modeling capability for the physical world significantly improved. 
This further indicates that Video-GPT has substantial room for improvement, 
as its pretraining can utilize almost all video data available on the internet without annotations of other modalities, similar to GPT.

\subsection{Downstream Task Evaluation}
We fine-tuned Video-GPT on downstream tasks, 
% Despite being trained primarily for generation tasks, 
% it achieved remarkable performance on understanding tasks as well. 
it demonstrates that our model acquired excellent prior knowledge and modal representations through large-scale self-supervised pretraining with videos. 
Unless specified, 
% we employ the Stage 3 pretraining Video-GPT  that completed Stage 3 in Tab. \ref{table: progressive_setting}.
we perform experiments based on the Stage 3 pretraining model in Tab. \ref{table: progressive_setting}.

\textbf{Class-to-Video Generation.}
As shown in Tab. \ref{table:c2v}, 
Video-GPT achieves the state-of-the-art performance on UCF-101 at high resolution, 
and we do not use the better 3D VAE, 
but 2D VAE, 
which means that our model still has a lot of room for improvement in terms of FVD. 
In addition, 
we test the Video-GPT trained from scratch on UCF-101, 
and the results show that our pretraining plays a very important role in the downstream task of class-to-video.
Qualitative results are presented in Fig. \ref{fig:ucf101_gen}.

\textbf{Text-to-Video Generation.}
We employ the same text encoder as SDv2 \cite{Rombach2021HighResolutionIS}.
We filter OpenVid-1M based on motion magnitude and aesthetic scores \cite{Lin2024OpenSoraPO} and obtain 0.3M pairs for training,
which is a relatively small dataset for text-to-video task. 
As shown in Fig. \ref{fig:T2V},
it demonstrates that our Video-GPT can achieve effective cross-modal generation with only a limited amount of text-video pairs.

\textbf{Image Animation.}
As shown in Fig. \ref{fig:image_animation}, 
we fine-tune Video-GPT 2K steps on three training sets with no more than 100 videos each, 
it showed amazing generalization ability on test cases outside the training set, 
fully demonstrating the powerful knowledge transfer ability of our Video-GPT.

\textbf{Video Classification.}
As shown in Tab. \ref{tab:video_classification},
Video-GPT linear probe on UCF-101 surpasses VideoMAEv2 \cite{Wang2023VideoMAEVS}, 
a pretrained model commonly used in the field of video understanding.

\textbf{Video Retrieval.}
As shown in Tab. \ref{tab:video_retrieval},
Video-GPT achieves impressive zero-shot performance on the MSR-VTT \cite{Xu2016MSRVTTAL} dataset and surpasses VideoClip \cite{Xu2021VideoCLIPCP} which was trained from scratch.

\textbf{Video Object Segmentation.}
We fine-tune Video-GPT 5K steps on the 1K subset of GetIn-1M.
As shown in Fig. \ref{fig:video_object_seg},,
the results show that the pretrained Video-GPT present good generalization performance when transferred to low level video understanding task.
\section{Conclusion}
\label{sec_conclusion}

We introduced Video-GPT, 
a concise yet powerful large video model that unites autoregressive modeling with diffusion through next clip diffusion. 
Treating each video clip as a word token allows the model to inherit the self-supervised property of GPT while retaining the sharp synthesis quality of diffusion. 
Pretraining on 70M unlabeled videos yields state-of-the-art accuracy on deterministic Physics-IQ Benchmark and uncertain Kinetics-600 forecasting, 
and fine-tuning transfers seamlessly to six diverse downstream tasks including generation and understanding. 
% Ablations confirm the benefit of longer training windows, historical guidance and scalable data. 
Future work will explore multi-modal pretraining, 
reinforcement-driven world interaction and further efficiency optimizations.

\bibliographystyle{abbrvnat}

\bibliography{reference}

%%%%%%%%%%%%%%%%%%%%%%%%%%%%%%%%%%%%%%%%%%%%%%%%%%%%%%%%%%%%
\newpage
\appendix

\section{More Implementation Details}
\label{sup:implementation}

\begin{table}[!htbp]
\centering
\begin{minipage}[tl]{0.4\linewidth}
    % \vspace{0pt}
    \raggedright
    \caption{\textbf{Pretraining Stage 1 setting.}
    }
    % \vspace{-0.2cm}
    \belowrulesep=0pt\aboverulesep=0pt
        \resizebox{\textwidth}{!}{
        \begin{tabular}{l|c}
        config & Panda-70M \\
        \Xhline{1.0pt}
        optimizer & AdamW \cite{Loshchilov2017DecoupledWD} \\ 
        optimizer momentum & $\beta_1, \beta_2{=}0.9, 0.95$  \\
        weight decay & 0.1 \\
        learning rate schedule & consistent \\
        learning rate & 1e-4 \\
        warmup steps \cite{Goyal2017AccurateLM} & 1000 \\
        total steps &  300000 \\
        input frame & 16 \\
        resulution & flexible \\
        longest side & 320 \\
        frame interval & 4 \\
        video clip & 16 \\
        computing unit & 320 \\
        \end{tabular}
    }
    \label{tab:pretrain_stage1}
\end{minipage}
\hfill
\begin{minipage}[tr]{0.48\linewidth}
    % \vspace{0pt}
    \raggedleft
    \caption{\textbf{Pretraining Stage 2 setting.}}
    % \vspace{-0.2cm}
    \belowrulesep=0pt\aboverulesep=0pt
        \resizebox{\textwidth}{!}{
        \begin{tabular}{l|c}
        config & Panda-70M \\
        \Xhline{1.0pt}
        optimizer & AdamW \cite{Loshchilov2017DecoupledWD} \\ 
        optimizer momentum & $\beta_1, \beta_2{=}0.9, 0.95$  \\
        weight decay & 0.1 \\
        learning rate schedule & consistent \\
        learning rate & 1e-4 \\
        warmup steps \cite{Goyal2017AccurateLM} & 1000 \\
        total steps &  25000 \\
        input frame & 48 \\
        resulution & flexible \\
        longest side & 320 \\
        frame interval & 4 \\
        video clip & $\sim \text{Uniform}\{2,3,\cdots,48\}$ \\
        computing unit & 320 \\
        \end{tabular}
    }
    % \vspace{0.00001cm}
    
    % \vspace{-0.3cm}
    \label{tab:pretrain_stage2}
\end{minipage}
% \vspace{-0.3cm}
\end{table}
\begin{table}[!htbp]
\centering
\begin{minipage}[tl]{0.48\linewidth}
    % \vspace{0pt}
    \raggedright
    \caption{\textbf{Pretraining Stage 3 setting.}
    }
    % \vspace{-0.2cm}
    \belowrulesep=0pt\aboverulesep=0pt
        \resizebox{\textwidth}{!}{
        \begin{tabular}{l|c}
        config & Panda-70M \\
        \Xhline{1.0pt}
        optimizer & AdamW \cite{Loshchilov2017DecoupledWD} \\ 
        optimizer momentum & $\beta_1, \beta_2{=}0.9, 0.95$  \\
        weight decay & 0.1 \\
        learning rate schedule & consistent \\
        learning rate & 1e-4 \\
        warmup steps \cite{Goyal2017AccurateLM} & 1000 \\
        total steps &  40000 \\
        input frame & 48 \\
        resulution & flexible \\
        longest side & 320 \\
        frame interval & $\sim \text{Uniform}\{4,5,\cdots,12\}$ \\
        video clip & $\sim \text{Uniform}\{2,3,\cdots,48\}$ \\
        computing unit & 320 \\
        \end{tabular}
    }
    \label{tab:pretrain_stage3}
\end{minipage}
\hfill
\begin{minipage}[tr]{0.48\linewidth}
    % \vspace{0pt}
    \raggedleft
    \caption{\textbf{Pretraining Stage 4 setting.}}
    % \vspace{-0.2cm}
    \belowrulesep=0pt\aboverulesep=0pt
        \resizebox{\textwidth}{!}{
        \begin{tabular}{l|c}
        config & Panda-70M \\
        \Xhline{1.0pt}
        optimizer & AdamW \cite{Loshchilov2017DecoupledWD} \\ 
        optimizer momentum & $\beta_1, \beta_2{=}0.9, 0.95$  \\
        weight decay & 0.1 \\
        learning rate schedule & consistent \\
        learning rate & 2e-5 \\
        warmup steps \cite{Goyal2017AccurateLM} & 1000 \\
        total steps &  20000 \\
        input frame & 80 \\
        resulution & flexible \\
        longest side & 320 \\
        frame interval & $\sim \text{Uniform}\{4,5,\cdots,12\}$ \\
        video clip & $\sim \text{Uniform}\{2,3,\cdots,80\}$ \\
        computing unit & 320 \\
        \end{tabular}
    }
    % \vspace{0.00001cm}
    
    % \vspace{-0.3cm}
    \label{tab:pretrain_stage4}
\end{minipage}
% \vspace{-0.3cm}
\end{table}
\begin{table}[!htbp]
\centering
\begin{minipage}[tl]{0.48\linewidth}
    % \vspace{0pt}
    \raggedright
    \caption{\textbf{Class to video fine-tuning on UCF-101 setting.}
    }
    % \vspace{-0.2cm}
    \belowrulesep=0pt\aboverulesep=0pt
        \resizebox{\textwidth}{!}{
        \begin{tabular}{l|c}
        config & UCF-101 \\
        \Xhline{1.0pt}
        optimizer & AdamW \cite{Loshchilov2017DecoupledWD} \\ 
        optimizer momentum & $\beta_1, \beta_2{=}0.9, 0.95$  \\
        weight decay & 0.1 \\
        learning rate schedule & consistent \\
        learning rate & 1e-4 \\
        warmup steps \cite{Goyal2017AccurateLM} & 1000 \\
        total steps &  165000 \\
        input frame & 16 \\
        resulution & 240$\times$320 \\
        longest side & 320 \\
        frame interval & 1 \\
        video clip & 1 \\
        computing unit & 64 \\
        \end{tabular}
    }
    \label{tab:finetune_c2v}
\end{minipage}
\hfill
\begin{minipage}[tr]{0.48\linewidth}
    % \vspace{0pt}
    \raggedleft
    \caption{\textbf{Text to video fine-tuning on OpenVid-0.3M setting.}}
    % \vspace{-0.2cm}
    \belowrulesep=0pt\aboverulesep=0pt
        \resizebox{\textwidth}{!}{
        \begin{tabular}{l|c}
        config & OpenVid-0.3M \\
        \Xhline{1.0pt}
        optimizer & AdamW \cite{Loshchilov2017DecoupledWD} \\ 
        optimizer momentum & $\beta_1, \beta_2{=}0.9, 0.95$  \\
        weight decay & 0.1 \\
        learning rate schedule & consistent \\
        learning rate & 1e-4 \\
        warmup steps \cite{Goyal2017AccurateLM} & 1000 \\
        total steps &  70000 \\
        input frame & 24 \\
        resulution & flexible \\
        longest side & 320 \\
        frame interval & 4 \\
        video clip & 1 \\
        computing unit & 128 \\
        \end{tabular}
    }
    % \vspace{0.00001cm}
    
    % \vspace{-0.3cm}
    \label{tab:finetune_t2v}
\end{minipage}
% \vspace{-0.3cm}
\end{table}
\begin{table}[!htbp]
\centering
\begin{minipage}[tl]{0.48\linewidth}
    % \vspace{0pt}
    \raggedright
    \caption{\textbf{Image animation fine-tuning on animation dataset setting.}
    }
    % \vspace{-0.2cm}
    \belowrulesep=0pt\aboverulesep=0pt
        \resizebox{\textwidth}{!}{
        \begin{tabular}{l|c}
        config & Anim. Dataset \\
        \Xhline{1.0pt}
        optimizer & AdamW \cite{Loshchilov2017DecoupledWD} \\ 
        optimizer momentum & $\beta_1, \beta_2{=}0.9, 0.95$  \\
        weight decay & 0.1 \\
        learning rate schedule & consistent \\
        learning rate & 1e-4 \\
        warmup steps \cite{Goyal2017AccurateLM} & 1000 \\
        total steps &  2000 \\
        input frame & 48 \\
        resulution & flexible \\
        longest side & 320 \\
        frame interval & flexible \\
        video clip & 2 \\
        computing unit & 8 \\
        \end{tabular}
    }
    \label{tab:finetune_image_anim}
\end{minipage}
\hfill
\begin{minipage}[tr]{0.48\linewidth}
    % \vspace{0pt}
    \raggedleft
    \caption{\textbf{Video classification fine-tuning on UCF-101 setting.}}
    % \vspace{-0.2cm}
    \belowrulesep=0pt\aboverulesep=0pt
        \resizebox{\textwidth}{!}{
        \begin{tabular}{l|c}
        config & UCF-101 \\
        \Xhline{1.0pt}
        optimizer & AdamW \cite{Loshchilov2017DecoupledWD} \\ 
        optimizer momentum & $\beta_1, \beta_2{=}0.9, 0.95$  \\
        weight decay & 0.1 \\
        learning rate schedule & consistent \\
        learning rate & 1e-4 \\
        warmup steps \cite{Goyal2017AccurateLM} & 1000 \\
        total steps &  10000 \\
        input frame & 16 \\
        resulution & 224$\times$224 \\
        longest side & 224 \\
        frame interval & 1 \\
        video clip & 1 \\
        computing unit & 8 \\
        \end{tabular}
    }
    % \vspace{0.00001cm}
    
    % \vspace{-0.3cm}
    \label{tab:finetune_video_classification}
\end{minipage}
% \vspace{-0.3cm}
\end{table}
\begin{table}[!htbp]
\centering
\begin{minipage}[tl]{0.48\linewidth}
    % \vspace{0pt}
    \raggedright
    \caption{\textbf{Video retrieval fine-tuning on Panda-70M.}
    }
    % \vspace{-0.2cm}
    \belowrulesep=0pt\aboverulesep=0pt
        \resizebox{\textwidth}{!}{
        \begin{tabular}{l|c}
        config & Panda-70M \\
        \Xhline{1.0pt}
        optimizer & AdamW \cite{Loshchilov2017DecoupledWD} \\ 
        optimizer momentum & $\beta_1, \beta_2{=}0.9, 0.95$  \\
        weight decay & 0.1 \\
        learning rate schedule & consistent \\
        learning rate & 1e-4 \\
        warmup steps \cite{Goyal2017AccurateLM} & 1000 \\
        total steps &  200000 \\
        input frame & 8 \\
        resulution & 224$\times$224 \\
        longest side & 224 \\
        frame interval & flexible \\
        video clip & 1 \\
        computing unit & 64 \\
        \end{tabular}
    }
    \label{tab:finetune_video_retrieval}
\end{minipage}
\hfill
\begin{minipage}[tr]{0.48\linewidth}
    % \vspace{0pt}
    \raggedleft
    \caption{\textbf{Video object segmenration fine-tuning on GetIn-1K setting.}}
    % \vspace{-0.2cm}
    \belowrulesep=0pt\aboverulesep=0pt
        \resizebox{\textwidth}{!}{
        \begin{tabular}{l|c}
        config & GetIn-1K \\
        \Xhline{1.0pt}
        optimizer & AdamW \cite{Loshchilov2017DecoupledWD} \\ 
        optimizer momentum & $\beta_1, \beta_2{=}0.9, 0.95$  \\
        weight decay & 0.1 \\
        learning rate schedule & consistent \\
        learning rate & 1e-4 \\
        warmup steps \cite{Goyal2017AccurateLM} & 1000 \\
        total steps &  5000 \\
        input frame & 48 \\
        resulution & flexible \\
        longest side & 224 \\
        frame interval & flexible \\
        video clip & 2 \\
        computing unit & 8 \\
        \end{tabular}
    }
    % \vspace{0.00001cm}
    
    % \vspace{-0.3cm}
    \label{tab:finetune_video_object_segmenration}
\end{minipage}
% \vspace{-0.3cm}
\end{table}

\section{More Visualization Results}
\label{sup:visualization}

\begin{figure}[!htbp]
    \centering
    % \vspace{-0.3cm}   
    \includegraphics[width=1\textwidth]{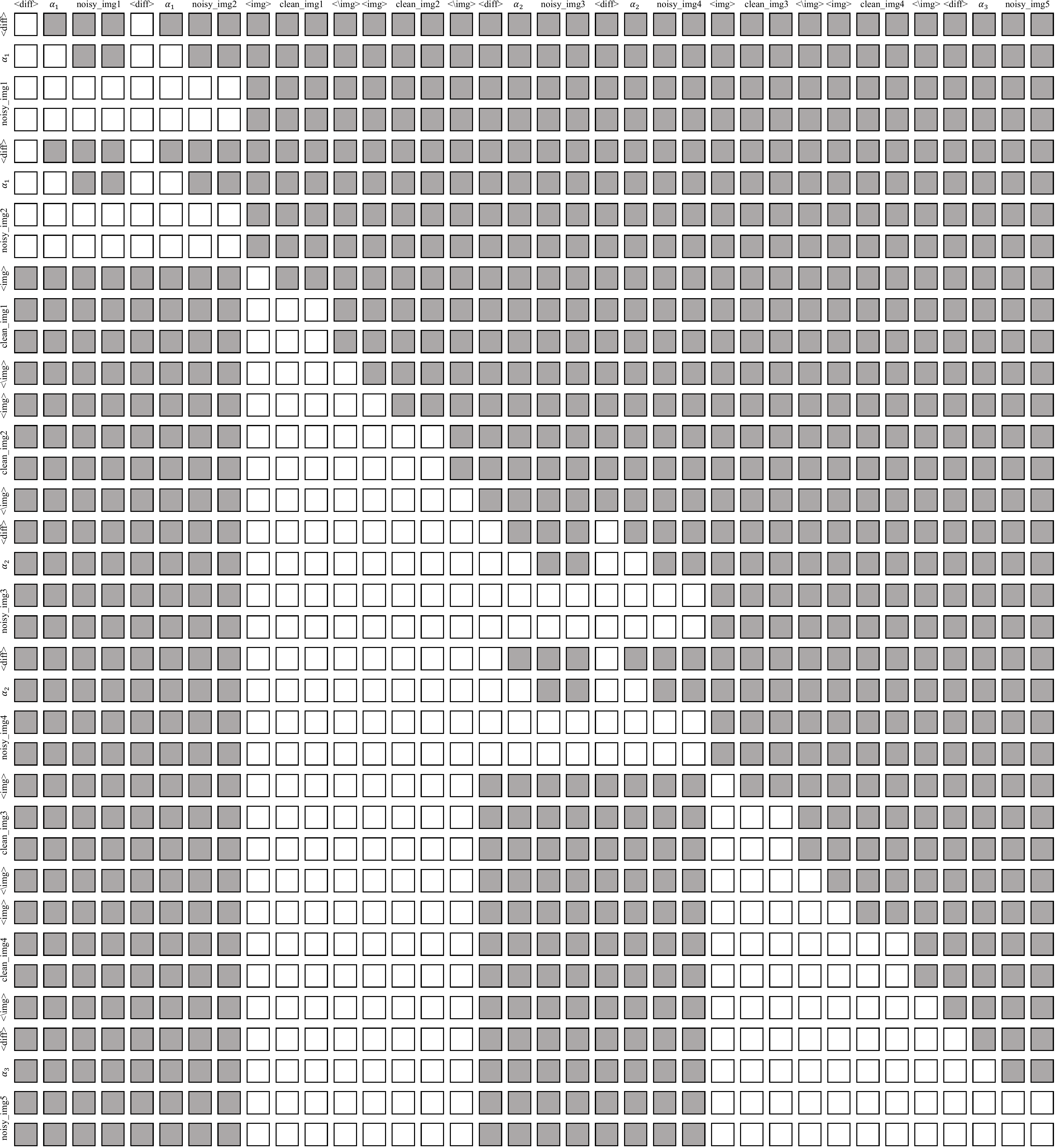}
    \vspace{-0.5cm}
    \caption{
    \textbf{The full attention mask of Fig. \ref{fig:model_pretrain} (a) training input.
    } 
    }
    \label{fig:full_mask}
    % \vspace{-0.1cm}
\end{figure}

The full mask of the input in Fig. \ref{fig:model_pretrain} (a) is shown in Fig. \ref{fig:full_mask}.

\begin{figure}[!htbp]
    \centering
    % \vspace{-0.3cm}   
    \includegraphics[width=1\textwidth]{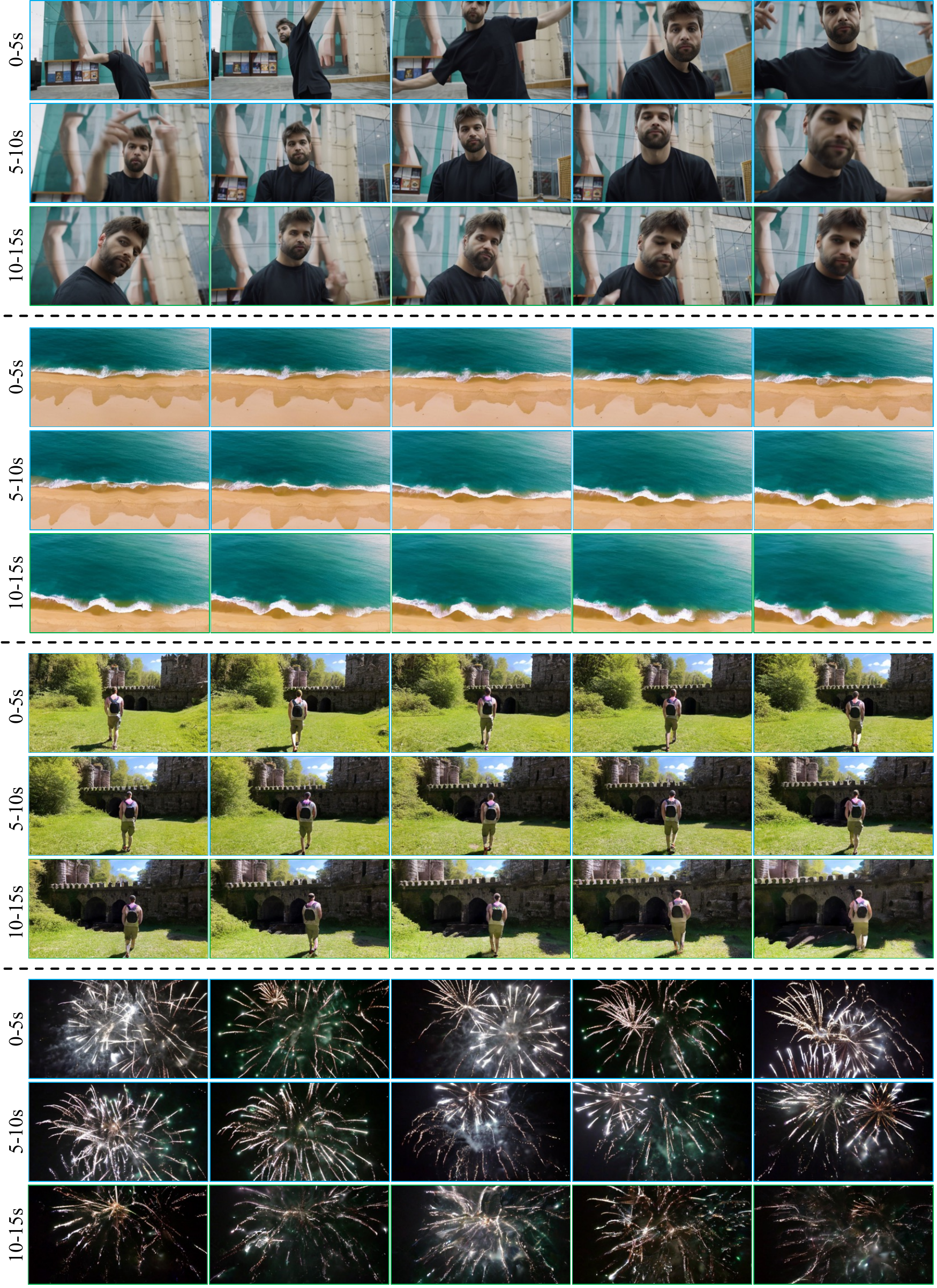}
    % \vspace{-0.5cm}
    \caption{
    \textbf{The long video zero-shot prediction result generated by Video-GPT.
    } 
    The frames in \textcolor[RGB]{79,173,234}{blue} box are condition frames.
    Frames in \textcolor[RGB]{79,173,91}{green} box are the generated prediction result.
    }
    \label{fig:long_vid_wide}
    % \vspace{-0.1cm}
\end{figure}

\begin{figure}[!htbp]
    \centering
    % \vspace{-0.3cm}   
    \includegraphics[width=0.85\textwidth]{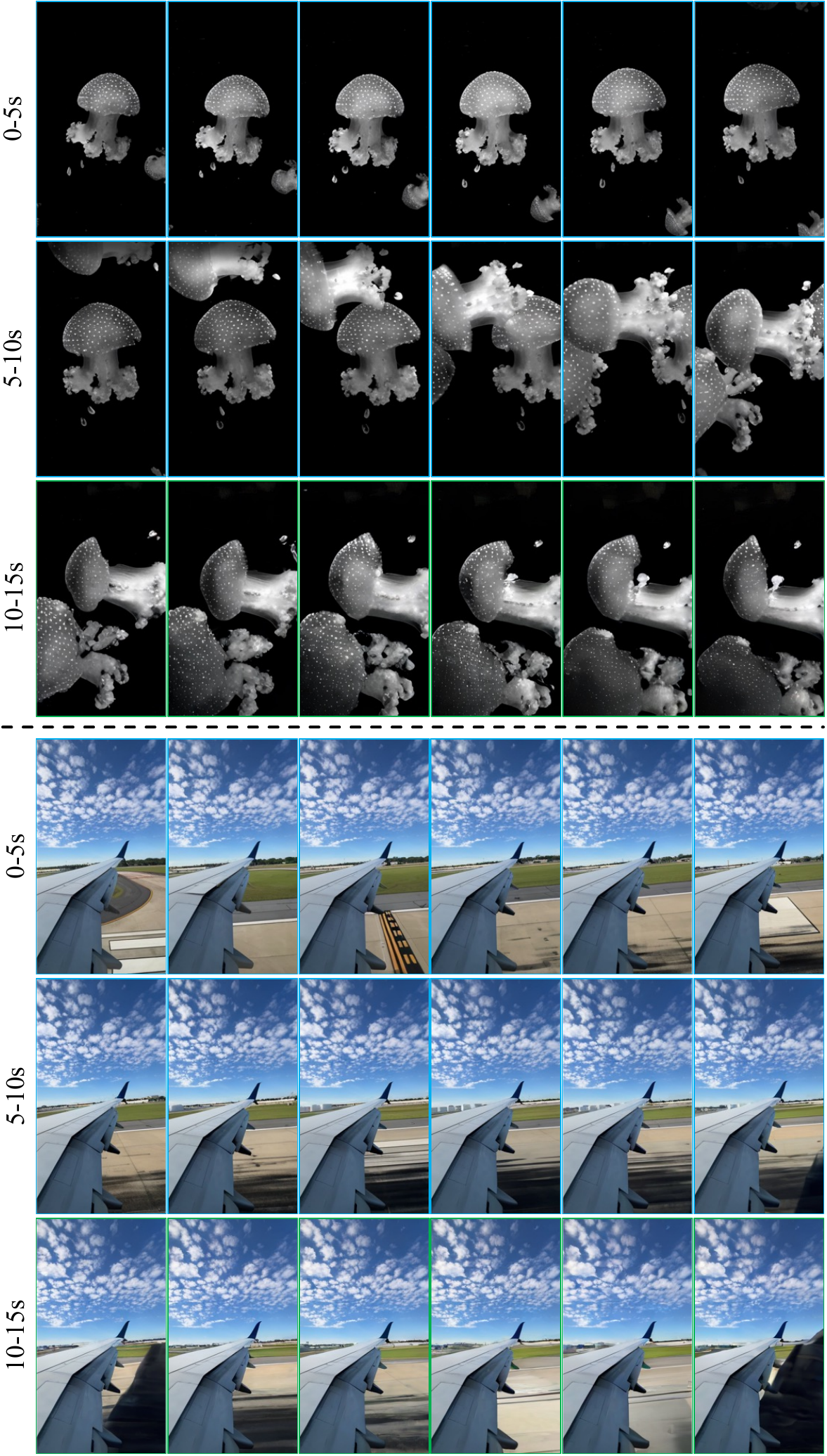}
    % \vspace{-0.5cm}
    \caption{
    \textbf{The long video zero-shot prediction result generated by Video-GPT.
    } 
    The frames in \textcolor[RGB]{79,173,234}{blue} box are condition frames.
    Frames in \textcolor[RGB]{79,173,91}{green} box are the generated prediction result.
    }
    \label{fig:long_vid_high}
    % \vspace{-0.1cm}
\end{figure}

We show the results of Video-GPT for zero-shot long video prediction in Fig. \ref{fig:long_vid_wide} and \ref{fig:long_vid_high}.

\section{Limitations}
\label{sup:limit}

Our model exhibits remarkable performance and generalization across diverse tasks, 
affirming the strength of the next clip diffusion paradigm. 
It is important to note, 
however, 
that our experiments are conducted with a moderately scaled model due to available computational resources. 
This choice, 
while pragmatic, 
has not detracted from the quality of our good results. 
Looking forward, 
further scaling may reveal additional performance gains. 
Nonetheless, 
the current configuration already sets a high benchmark, 
underscoring both the efficacy and robustness of our approach in video generative modeling.

\end{document}